\documentclass[sigconf]{acmart}

\setcopyright{none}
\renewcommand\footnotetextcopyrightpermission[1]{}

\AtBeginDocument{%
  \providecommand\BibTeX{{%
    Bib\TeX}}}

\usepackage{enumitem}
\usepackage{algorithm}
\usepackage{algorithmic}

\usepackage{hyperref}       % hyperlinks
\usepackage{url}            % simple URL typesetting
\usepackage{booktabs}       % professional-quality tables
\usepackage{amsfonts}       % blackboard math symbols
\usepackage{nicefrac}       % compact symbols for 1/2, etc.
\usepackage{xcolor}         % colors
\usepackage{amsmath}
\usepackage{makecell} % 添加在导言区以启用 \makecell

\usepackage{multirow}
\usepackage{soul}
\usepackage{subcaption}
\usepackage{bm}
\usepackage{wasysym}

\def\BibTeX{{\rm B\kern-.05em{\sc i\kern-.025em b}\kern-.08em
    T\kern-.1667em\lower.7ex\hbox{E}\kern-.125emX}}
\begin{document}

\title{Dataforge: Agentic Platform for Autonomous Data Engineering}

\author{Xinyuan Wang}
\affiliation{%
  \institution{Arizona State University}
  \city{Tempe}
  \state{Arizona}
  \country{USA}}
\email{xwang735@asu.edu}

\author{Hongyu Cao}
\affiliation{%
  \institution{Arizona State University}
  \city{Tempe}
  \state{Arizona}
  \country{USA}}
\email{hongyuca@asu.edu}

\author{Kunpeng Liu}
\affiliation{%
  \institution{Clemson University}
  \city{Clemson}
  \state{South Carolina}
  \country{USA}}
\email{kunpenl@clemson.edu}

\author{Yanjie Fu}
\affiliation{%
  \institution{Arizona State University}
  \city{Tempe}
  \state{Arizona}
  \country{USA}}
\email{yanjie.fu@asu.edu}

\renewcommand{\shortauthors}{Trovato et al.}

\begin{abstract}
The growing demand for artificial intelligence (AI) applications in materials discovery, molecular modeling, and climate science has made data preparation a critical but labor-intensive bottleneck.
Raw data from diverse sources must be cleaned, normalized, and transformed to become AI-ready, where effective feature transformation and selection are essential for robust learning.
We present \textsc{Dataforge}, an LLM-powered agentic data engineering platform for tabular data that is \textit{automatic}, \textit{safe}, and \textit{non-expert friendly}.
It autonomously performs data cleaning and iteratively optimizes feature operations under a budgeted feedback loop with automatic stopping.
Across tabular benchmarks, it achieves the best overall downstream performance; ablations further confirm the roles of routing/iterative refinement and grounding in accuracy and reliability.
\textsc{Dataforge} demonstrates a practical path toward autonomous data agents that transform raw data \emph{from data to better data}.
\end{abstract}

%    <concept>
%        <concept_id>10002951.10003227.10003351</concept_id>
%        <concept_desc>Information systems~Data mining</concept_desc>
%        <concept_significance>500</concept_significance>
%        </concept>
%    <concept>
%        <concept_id>10010147.10010178</concept_id>
%        <concept_desc>Computing methodologies~Artificial intelligence</concept_desc>
%        <concept_significance>500</concept_significance>
%        </concept>
%  </ccs2012>

% \ccsdesc[500]{Information systems~Data mining}
% \ccsdesc[500]{Computing methodologies~Artificial intelligence}

\keywords{Data Agent, LLMs, Data-Centric Agentic AI}


\maketitle

\section{Introduction}
%As AI technologies continue to expand into domains such as materials discovery, molecular modeling, and climate simulation, the ability to prepare and transform data has become critical~\cite{yu2025ai}.  In many AI4Science scenarios, raw data collected from heterogeneous sources must undergo extensive processing before it becomes usable by AI models.  Cleaning, integration, and normalization are required to ensure structural and semantic consistency so that downstream models can function properly~\cite{fernandes2023data}, while operations such as feature transformation and selection aim to make even simple AI models perform more effectively~\cite{wang2025towards}, improving both training stability and inference efficiency.  Despite their importance, these data-centric tasks remain repetitive, time-consuming, and error-prone, often requiring domain experts to manually design and validate the processing pipeline.
Modern AI applications increasingly rely on AI-ready data that are clean, consistent, and semantically aligned with the objectives of the model. In fields such as materials discovery, molecular design, and climate modeling, AI-ready data determine whether models can achieve reliable prediction, efficient training, and meaningful scientific insight~\cite{yu2025ai}. However, transforming raw data into such an AI-ready form remains a persistent challenge~\cite{wang2025towards,ying2025survey}. Raw datasets are often heterogeneous, incomplete, and noisy, collected from instruments or repositories with inconsistent formats and semantics. Converting them into usable inputs requires extensive cleaning, normalization, and feature transformation, which are still largely manual, error-prone, and dependent on domain expertise. This growing mismatch between the rapid progress of AI models and the slow and fragmented data preparation process exposes a fundamental data engineering gap that increasingly limits the scalability and reliability of AI applications.

%Meanwhile, the rapid advances in large language models (LLMs) have brought a new opportunity.  LLMs not only encode vast domain knowledge from texts but also demonstrate strong reasoning and instruction-following abilities, which can be leveraged to interpret, plan, and execute data-related operations.  This convergence of AI and data processing gives rise to the concept of \textbf{Data Agents}, where autonomous agents~\cite{durante2024agent} can understand a dataset, reason about transformations, and execute operations toward improving data quality and utility. 

The rapid progress of large language models (LLMs) has opened a new path toward autonomous data engineering. LLMs not only capture extensive domain knowledge from large-scale text corpora but also exhibit emerging abilities to interpret data schemas, infer transformation logic, and generate executable operations—tasks traditionally reserved for human experts~\cite{xie2025text}. Although many recent works have explored using LLMs for data-centric tasks such as LLM-based feature engineering and transformation~\cite{han2024large,hollmann2023large,horn2019autofeat,zhang2023openfe,gong2025evolutionary}, they remain limited in several aspects:
(1) \textbf{Lack of full automation:} Most approaches still require user intervention, manual code review, or step-by-step prompting to complete a data pipeline.
(2) \textbf{Limited reliability:} LLMs may generate invalid or unsafe operations without schema verification or grounding checks, leading to broken workflows.
(3) \textbf{Expert dependency:} Many systems assume the user understands LLM prompts, data schemas, or tool APIs, which prevents non-expert adoption.

% \vspace{-0.1cm}
Newer LLMs now demonstrate emergent agentic behaviors: the ability to reason, plan, and act through iterative interaction with tools and environments. This shift marks an agentic evolution in how intelligence is deployed: systems no longer just generate outputs for a descriptive query, but autonomously understand tasks, reason, plan, ground, execute actions, and adapt through feedback. Extending this paradigm to the data domain gives rise to the concept of Data Agents: autonomous systems that can understand datasets, reason about transformations, and execute data-centric operations to improve quality, consistency, and utility.

\begin{figure}[htbp]
  \centering
  \includegraphics[width=\linewidth]{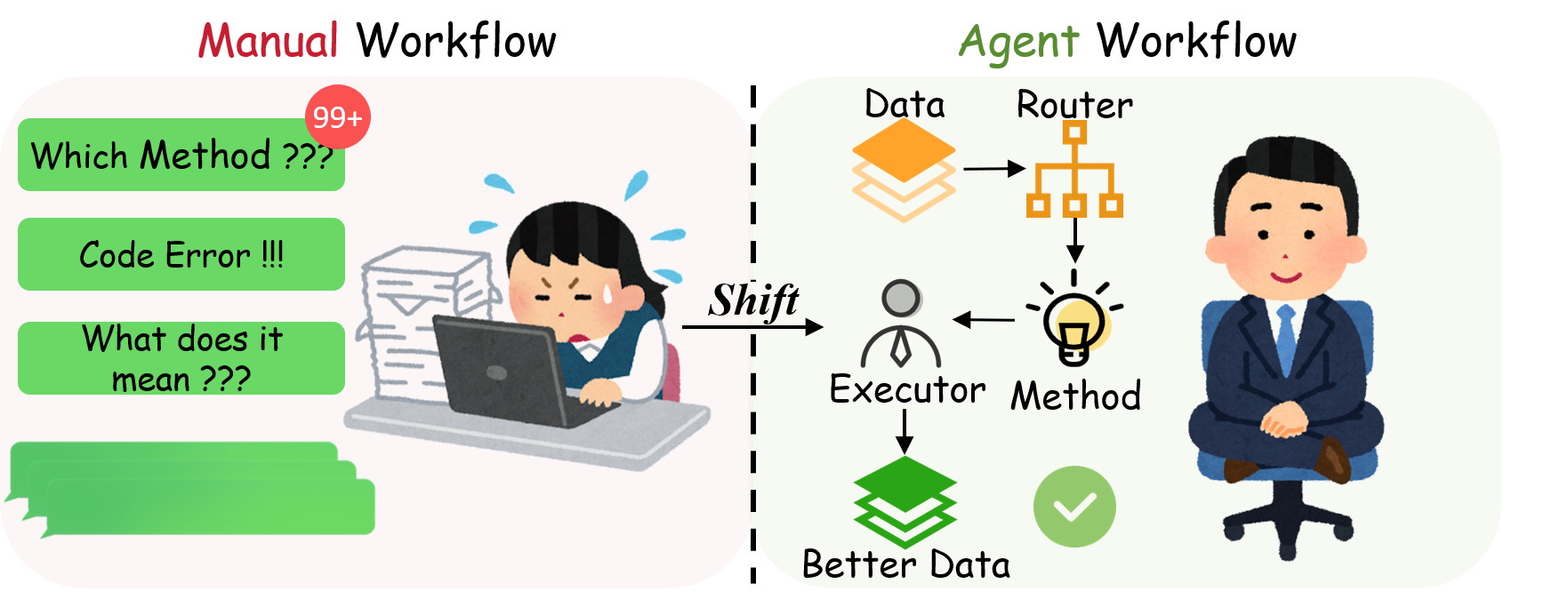}
  \caption{
  Conceptual comparison between a traditional manual workflow and the agentic workflow.}
  \label{fig:intro_long}
\end{figure}

% \vspace{-0.1cm}
To this end, we develop the \textsc{Dataforge}, an autonomous data agent platform specialized for tabular data that embodies three core design goals: 1) \textbf{Automatic:} performs end-to-end data transformation without manual intervention; 2) \textbf{Safe:} ensures workflow reliability via routing, grounding, and execution validation; 3) \textbf{Non-expert Friendly:} allows users with minimal technical background to improve their data by simply specifying task and data types.
The \textsc{Dataforge} is implemented based on the planning–grounding–execution principles~\cite{fu2025autonomous} as a deployable system for tabular data. The data agent includes a sequential set of stages of data cleaning, hierarchical routing, grounding validation, and dual-loop optimization. 
In particular, the hierarchical routing mechanism determines both the high-level task family and the low-level feature actions; each action is checked before execution to ensure safety, and results are evaluated to guide iterative improvement; the dual feedback loops ensure that each decision is validated before execution and that performance improvements are optimized through experience replay. 

Our empirical study demonstrates that \textsc{Dataforge} provides strong and robust gains across diverse tabular datasets: it achieves the best overall downstream utility on six classification and three regression benchmarks compared with classical, RL-based, and LLM-based approaches.
Ablation results further verify that hierarchical routing and evaluation-driven refinement are key to performance, while grounding substantially improves executability by reducing invalid actions and repeated retries.
It also avoids RL-style training cost and achieves strong performance with low call complexity, supporting its practicality as an automated data engineering tool.

\textbf{Our Contribution:}
We present \textsc{Dataforge}, an LLM-powered agentic data engineering platform for tabular data that is \emph{automatic}, \emph{safe}, and \emph{non-expert friendly}, iteratively orchestrating feature operations with automatic stopping to produce AI-ready data.

\section{Preliminaries}

\subsection{Problem Formulation}
We consider automated data preparation and feature engineering for tabular datasets.
Let $\mathcal{D}=\{(x_i, y_i)\}_{i=1}^{n}$ denote a dataset with a feature table $X\in\mathbb{R}^{n\times d}$ (mixed numeric/categorical/time columns) and an optional target column $y$.
Given $\mathcal{D}$ and a task specification $\mathcal{T}$ (e.g., classification or regression, evaluation metric, and optional constraints), our goal is to construct a transformation pipeline $\pi$ that maps the raw dataset to an AI-ready representation:
\[
\pi: \mathcal{D} \rightarrow \mathcal{D}'.
\]

A pipeline $\pi$ is a sequence of executable feature operations $\pi=[a_1,\ldots,a_K]$, where each action $a_k$ is selected from a tool library $\mathcal{A}$ (e.g., missing-value imputation, encoding, scaling, feature generation/selection).
Applying $\pi$ produces a transformed dataset $\mathcal{D}'$ and an execution log.
We optimize $\pi$ under a user-defined budget $\mathcal{B}$ (e.g., time limit, maximum number of actions) to improve downstream performance while preserving correctness and safety:
\[
\max_{\pi} \; \mathrm{Perf}(\mathcal{D}', \mathcal{T}) \quad
\text{s.t.}\quad \mathrm{Safe}(\pi,\mathcal{D})=1,\;\; \mathrm{Cost}(\pi)\le \mathcal{B}.
\]

Here, $\mathrm{Perf}(\cdot)$ is measured by a standard metric (e.g., F1 for classification, 1-RAE for regression) using a fixed evaluation protocol, and $\mathrm{Safe}(\cdot)$ indicates whether the pipeline satisfies grounding checks.

\subsection{Platform Value}

\textbf{End-to-end automation:}
Given a dataset and a minimal task specification, \textsc{Dataforge} automatically profiles data, proposes feature operations, validates them, executes transformations, and iteratively improves data utility under a budget.
\textbf{Reliability and safety:}
Unlike prompt-only approaches, \textsc{Dataforge} enforces action grounding and guardrails to prevent invalid or unsafe transformations and to ensure stable execution.
\textbf{Benchmarkable and reproducible outputs.}
\textsc{Dataforge} exports a replayable pipeline together with structured logs and summary reports.
\textbf{Non-expert usability:}
\textsc{Dataforge} exposes task-level controls rather than requiring users to craft prompts or write feature code, lowering the barrier for non-expert practitioners.

\subsection{System I/O Contract}
\textbf{Inputs:} (i) a tabular dataset (CSV/Parquet or a database table), (ii) an optional target column name, (iii) a task specification $\mathcal{T}$ including task type and evaluation metric, and (iv) optional constraints $\mathcal{B}$ such as time budget, maximum number of actions, and minimum improvement threshold for early stopping.

\textbf{Outputs:} (i) a transformed dataset $\mathcal{D}'$ (AI-ready table), (ii) a replayable pipeline $\pi$ that specifies the ordered list of feature operations with parameters, (iii) execution logs that record dataset diffs and validation outcomes for each step, and (iv) a summary report that explains key transformations and the measured utility improvements.

\textbf{Failure handling:}
If an action cannot be validated or executed, \textsc{Dataforge} returns a failure reason (e.g., schema mismatch, type violation, suspected leakage, or runtime error) and either (a) replans with an alternative action, or (b) terminates with the best valid pipeline found so far under the budget.

\subsection{Important Concepts}
\textbf{AI-ready data:}
We use \emph{AI-ready} to describe tabular data that are schema-consistent, type-valid, and suit to model training (e.g., reduced missingness and improved downstream performance).

\textbf{Operation Set:} To refine the feature space, we need to apply mathematical operations to existing features to generate new informative features. All operations are collected in an operation set, denoted by $\mathcal{O}$. These operations can be classified as unary and binary operations. The unary operations such as ``\texttt{square}'', ``\texttt{exp}'', ``\texttt{log}'', etc. The binary operations such as ``\texttt{plus}'', ``\texttt{multiply}'', ``\texttt{minus}'', etc.

\textbf{Feature Transformation Sequence:} Assuming a dataset $\mathcal{D}=\{(\mathbf{x}^{(i)}, y^{(i)})\}_{i=1}^{N}$ includes the original feature set $\mathcal{F}_0=\{f_1,\ldots,f_K\}$ and predictive targets $y$. We transform the existing features using mathematical compositions $\tau$ consisting of feature ID tokens and operations to generate new and informative features. $K$ compositions are adopted to refine $\mathcal{F}_0$ to a better feature space $\tilde{\mathcal{F}_0}=\{\tilde{f}_{1},\ldots,\tilde{f_K}\}$. The collection of the $K$ compositions refers to the feature transformation sequence, which is denoted by $\Gamma = [\tau_{1},\cdots,\tau_{K}]$.

\textbf{Postfix Expressions:} 
The transformation sequence should be in a computable and machine-learnable format. The original infix representation has issues like redundancy, semantic sparsity, a high likelihood of illegal transformations, and an overly large search space.
We introduce postfix expressions to solve these problems. 
Postfix expressions don't need many brackets to determine calculation priority. Scanning from left to right suffices to reconstruct the corresponding sequence, greatly reducing sequence-modeling difficulty and computational cost.
They also reduce the ambiguity of the transformation sequence.
Most importantly, it reduces the search space from exponential to a finite set $|C| = |\mathcal{O}| + |\mathcal{F}_0|D + 3$. Here, $|\mathcal{O}|$ represents the operation set size, $|\mathcal{F}_0|$ is the original feature set dimension, $D$ is feature numbers, and $3$ refers to start tokens $<SOS>$, separation token $<SEP>$, and end token $<EOS>$.

% \subsection{Problem Statement}
% We aim to develop a generative AI system that generates a feature transformation sequence given a tasking dataset. Formally, Let a tabular dataset be $\mathcal{D}=\{(\mathbf{x}^{(i)}, y^{(i)})\}_{i=1}^{N}$ with original feature set $\mathcal{F}_0=\{f_1,\ldots,f_d\}$.
% A feature transformation step applies operators $\mathsf{op}$ from the operation set $\mathcal{O}$ on the original features to generate combinations for new features.
% The goal is to find the optimal feature transformation sequence $\Gamma^{*}$ that maximizes the downstream ML model $\mathcal{M}$'s performance (i.e.,  balance among accuracy, validity, and stability) on the transformed feature set:
% \begin{equation}
% \Gamma^{*} = \underset{\Gamma}{\operatorname{argmax}} \mathcal{A}(\mathcal{M}(\text{Transform}(\mathcal{F}_0, \Gamma)), y)
% \end{equation}
% where $\text{Transform}(\mathcal{F}_0, \Gamma)$ transforms the original feature set $\mathcal{F}_0$ using $\Gamma$, and $\mathcal{A}$ is $\mathcal{M}$'s downstream performance metric.

\begin{figure*}[h]
  \centering
  \includegraphics[width=0.95\linewidth]{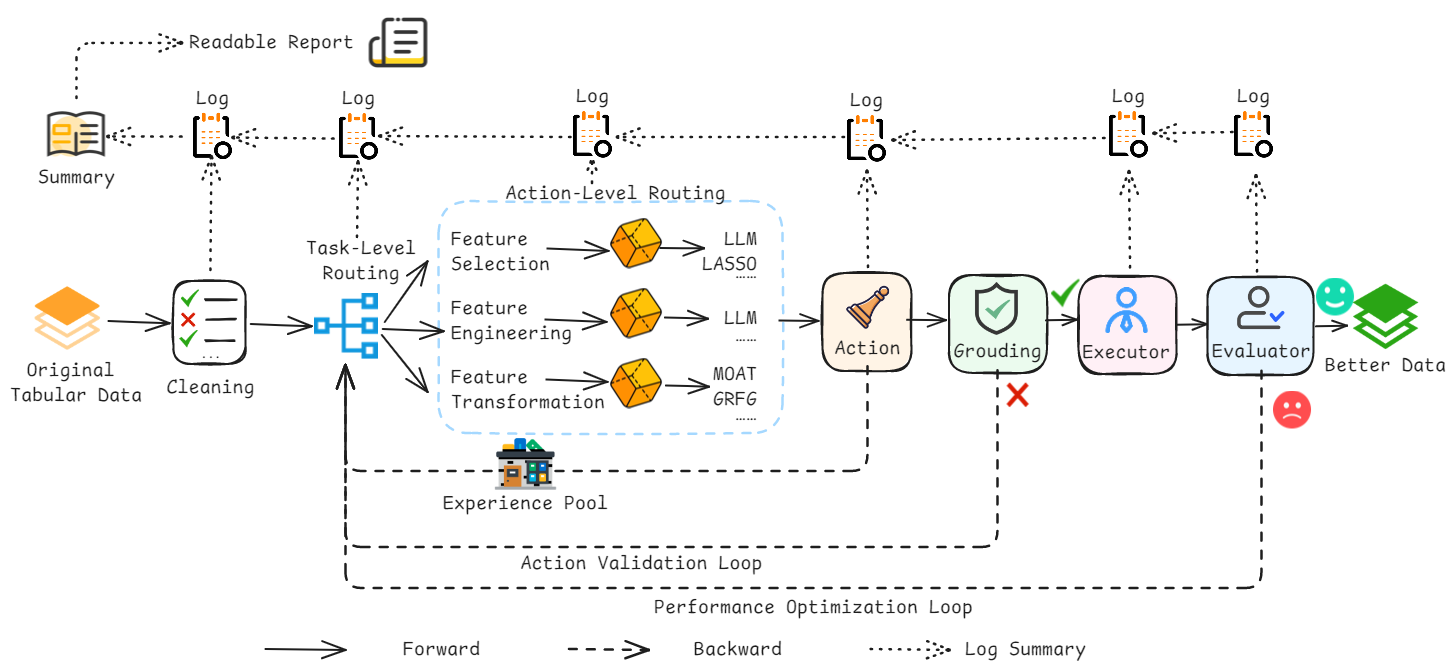}
    % \vspace{-0.2cm}
  \caption{The framework of the \textsc{Dataforge} system.}
  \label{fig:workflow}
  % \vspace{-0.5cm}
\end{figure*}

\section{\textsc{Dataforge}: Agentic Data Engineering}

% \vspace{-0.cm}
\subsection{Overview of the \textsc{Dataforge}}
% \vspace{-0.1cm}

% Fig.~\ref{fig:workflow} shows that Dataforge aims to automate the end-to-end data preparation process—from raw input to adaptive refinement—through a structured six-stage pipeline.  
% (1) \emph{Input and Cleaning:} Raw data is standardized via schema alignment, missing-value handling, and type correction under preset modes (\emph{light}, \emph{aggressive}, or \emph{time-series}).  
% (2) \emph{Task-Level Routing:} A rule-based router identifies the task type (e.g., classification, clustering, regression) and dispatches data to the corresponding action-level LLM router.  
% (3) \emph{Action-Level Routing:} Within each branch, an LLM-based planner selects, transforms, or generates features based on metadata and experience.  
% (4) \emph{Grounding and Execution:} Actions are validated for schema and logic safety, then executed through modular Python or SQL engines.  
% (5) \emph{Evaluation and Feedback:} An evaluator computes task-specific metrics (e.g., F1, AUC, RMSE) and maintains two feedback loops to regulate the process, in which the inner loop validates actions to prevent workflow failure. The outer loop optimizes performance across iterations. 
% (6) \emph{Experience Replay and Summary:} The system logs all operations as experiences to replay for continual learning and automatically generate reports for traceable provenance.

\textbf{Figure.~\ref{fig:workflow}} overviews \textsc{Dataforge}, which automates tabular data preparation and feature engineering via a structured, iterative pipeline.
(1) \emph{Input and Cleaning:} the input table is standardized through basic cleaning routines (schema/type normalization and missing-value handling) under preset modes (\emph{light}, \emph{aggressive}, or \emph{time-series}).
(2) \emph{Task-Level Routing:} a lightweight router infers the learning setting (e.g., classification, regression, or unsupervised) and activates the corresponding feature-action space.
(3) \emph{Action-Level Planning:} conditioned on dataset profiles and past trajectories, an LLM planner proposes candidate feature operations (selection, transformation, and generation) from a tool library.
(4) \emph{Grounding and Execution:} each proposed action is validated by grounding checks (e.g., schema/type compatibility and logical constraints) before being executed by the backend.
(5) \emph{Evaluation and Feedback:} the evaluator measures task-specific metric (e.g., F1/AUC/RMSE) and drives two feedback loops: an inner loop rejects unsafe or invalid actions, while an outer loop performs iterative re-planning toward improved performance under a budget.
(6) \emph{Experience and Reporting:} the system records action trajectories, validation outcomes, and metric progression, enabling experience-guided planning and generating a human-readable report for traceability.

%\subsection{Hierarchical Routing} To achieve efficient and safe decision-making, Data Agent employs a two-level hierarchical routing strategy that separates task-level and action-level reasoning.  This design decomposes complex decision-making into lightweight components, reducing both computation and uncertainty.  At the \textbf{I-Level (Task-Level Routing)}, a rule-based router quickly classifies the incoming dataset into task families such as \textit{classification}, \textit{regression}, or \textit{unsupervised learning}.  This layer uses predefined heuristics and metadata (e.g., data type, label structure, and feature distribution) to determine the appropriate processing branch.  Because it relies on simple rules rather than large models, it ensures fast response and stable behavior across diverse datasets. At the \textbf{II-Level (Action-Level Routing)}, a lightweight LLM-based module further selects and plans the most suitable feature-level actions within the chosen branch.  For example, given a classification dataset, the router may choose between feature selection, transformation, or generation actions.  By reducing the candidate space at each level, the system minimizes error propagation and computational overhead while maintaining flexibility.  This hierarchical approach not only accelerates processing but also improves safety, as each router operates within a smaller, well-defined action space, avoiding invalid or high-risk operations.

% \vspace{-0.2cm}
\subsection{Hierarchical Routing}
% \vspace{-0.1cm}
% After data cleaning, to enable efficient and reliable decision-making, we adopt a hierarchical routing architecture, including task-level and action-level reasoning. 
% At the \underline{\emph{task-level routing}}, a rule-based router quickly identifies the task type: \emph{classification}, \emph{regression}, or \emph{unsupervised learning}, based on table schema metadata, such as data types, label structures, and feature distribution. Such a lightweight router relies on deterministic heuristics, instead of large language models, thus enable fast and reliable responses across diverse datasets.
% At the \underline{\emph{action-level routing}}, a compact LLM-based planner refines the decision by selecting and planning the most suitable feature-level actions, such as, different ordered combinations of feature selection, transformation, or generation, under the identified task (e.g., a classification dataset). 
% Since each router operates within a smaller, well-defined action space, this hierarchical routing approach not only accelerates processing but also avoid invalid or high-risk operations.

After data cleaning, \textsc{Dataforge} adopts a hierarchical routing strategy~\cite{wang2025mixllm,bai2025learning} to make decisions more accurate and efficient.
Instead of directly searching over a large space of feature operations, we first perform a coarse-grained routing to a \emph{feature-operation family}, and then conduct fine-grained planning within the selected family.

\textbf{Task-level (coarse) routing:}
Here ``task'' refers to the \emph{data-engineering sub-task} at the current iteration, i.e., choosing among three high-level operation families: \emph{feature selection}, \emph{feature engineering}, and \emph{feature transformation} (\textbf{Figure.~\ref{fig:workflow}}).
The router is rule-based and uses lightweight dataset profiles (e.g., feature dimensionality, type composition, missingness, cardinality, and simple distributional statistics) to decide which family is most promising to apply next.
Intuitively, selection is preferred when the dataset exhibits strong redundancy or high dimensionality; transformation is preferred when scale/shape issues (e.g., heavy tails, outliers, or heterogeneous feature scales) dominate; and engineering is preferred when informative interactions or structured patterns are likely to exist.
This coarse routing restricts subsequent planning to a smaller and more relevant candidate set, reducing confusion across operation types.

\textbf{Action-level (fine) planning:}
Conditioned on the chosen family, an LLM-based planner selects concrete actions (e.g., specific selection criteria, transformation functions, or feature-generation methods) from the corresponding sub-library.
By operating within a well-defined action subspace, the planner produces more consistent candidates and avoids irrelevant proposals, which improves routing accuracy and reduces unnecessary trials.

\textbf{Why hierarchical routing:}
Let $\mathcal{A}$ denote the full tool library and $\mathcal{A}^{sel}$, $\mathcal{A}^{eng}$, $\mathcal{A}^{trans}$ denote the family-specific subsets.
Hierarchical routing first chooses one subset and then searches within it, effectively shrinking the candidate space at each step.
In our iterative setting, the outer optimization loop can revise the coarse choice in later iterations, enabling the system to alternate between families as needed.

\subsection{Feature Engineering Tool Library}
\textsc{Dataforge} organizes its feature operations into a family-structured tool library: \emph{feature selection}, \emph{feature transformation}, and \emph{feature engineering}.
Each tool corresponds to a parameterized feature action that can be proposed by the planner.

\textbf{Feature selection:}
This family targets redundancy reduction and robustness by selecting informative subsets of columns.
It includes standard filter and embedded operators (e.g., L1/LASSO, and tree-based importance), as well as group-wise selection primitives when features exhibit correlated blocks.
Beyond classical operators, the library includes policy-based selectors inspired by reinforcement-learning-based selection~\cite{liu2021automated, fan2021interactive, zhao2020simplifying, wang2025knockoff}, which propose candidate subsets under budgeted evaluations.

\textbf{Feature transformation:}
This family transforms existing features to improve numerical conditioning, handle scale and distribution mismatch, and mitigate noise/outliers, while preserving the original information content.
We include common transformations such as normalization/standardization, power/log transforms, clipping for outliers, and discretization/binning.
In addition to these hand-designed operators, the library can incorporate learned transformation modules that produce compact transformation programs or representations, leveraging reinforcement- or generation-based approaches to search over continuous or structured transformation spaces~\cite{wang2023reinforcement, ying2024unsupervised, gong2025unsupervised, wang2022group}.

\textbf{Feature engineering:}
This family constructs new features to expand representational capacity beyond the raw columns.
We support widely used engineering primitives, including interaction/cross features (e.g., pairwise products and conjunctions), logical compositions, and group-wise aggregation patterns, which are effective when predictive signals emerge from multi-column interactions.
For structured domains, the library can further incorporate domain-aware generators such as categorical hashing representations with hierarchical crossing and topology-aware feature space reconstruction for graph-structured attributes~\cite{ying2025topology, ying2023self}, enabling systematic feature creation when hand-crafted templates are insufficient.

% \vspace{-0.2cm}
\subsection{Dual Feedback Loops}
% \vspace{-0.1cm}
% We develop two collaborative feedback loops to transform the static workflow into an adaptive, self-correcting process, in order to achieve autonomy and continual refinement.
% %
% \noindent\underline{\emph{1) Action Validation Loop for Safety:}} This feddback loop is to ground actions to ensure operational safety before execution.   Each planned action is first grounded through schema alignment, type checking, and logical consistency tests, such as, detecting divisions by zero or invalid type conversions.  Only actions that pass validation proceed to execution so as to prevent runtime errors and maintaining workflow integrity.  
% %
% \noindent\underline{\emph{2) Performance Optimization Loop for Adaptiveness:}} 
% This feedback loop focuses on post-execution improvement and enabling the Dataforge to converge toward optimal behavior without human supervision. In particular, after each valid action runs, an evaluator measures task-specific metrics (e.g., F1, AUC, RMSE). If performance improves, the system updates its optimal configuration; otherwise, it triggers re-planning of new data engineering actions.  
% %
% The two loops establish a closed-cycle control mechanism for both grounded safety and self-optimizing adaptivity.

We develop two collaborative feedback loops that turn the workflow into an adaptive, self-correcting process.
The first loop ensures that every proposed action is \emph{executable} (grounding for safety), while the second loop guides the system toward \emph{better downstream performance} through iterative evaluation and re-planning.

\textbf{Action Validation Loop for Safety:} 
Given a candidate action proposed by the planner, \textsc{Dataforge} performs grounding checks before execution.
These checks validate (i) \emph{schema compatibility} (target columns exist and the action's output schema is well-formed), (ii) \emph{type consistency} (e.g., numeric operators are not applied to categorical strings; encoders receive valid category sets), and (iii) \emph{logical feasibility} (e.g., avoiding division by zero, invalid logarithms, or degenerate transformations that collapse a column to constants).
If any check fails, the action is rejected with an explicit failure reason, and the planner is prompted to revise the proposal.
Only grounded actions proceed to execution, which significantly reduces runtime errors and prevents invalid pipelines from contaminating subsequent iterations.

\textbf{Performance Optimization Loop for Adaptiveness:} 
After a grounded action is executed, the evaluator measures downstream performance under the given task specification (e.g., F1 for classification and 1-RAE for regression).
\textsc{Dataforge} maintains a \emph{best-so-far} state that records the current best pipeline and its evaluation score.
If the new action yields improvement beyond a small tolerance, the best state is updated and the action trajectory is stored as a positive experience; otherwise, the system rolls back to the best-so-far pipeline and triggers re-planning to explore alternative actions.
This loop enables \textsc{Dataforge} to iteratively search over feature operations without human intervention, while remaining robust to noisy or non-monotonic improvements.

\textbf{Stopping and budget control:}
Both loops are governed by user-defined budgets and early-stop criteria (e.g., maximum actions, minimum improvement, and time budget).
The optimization terminates when the budget is exhausted or when no meaningful improvement is observed for several iterations, returning the best validated pipeline and its transformed dataset.
Overall, the two loops form a closed-cycle mechanism that couples \emph{safe executability} with \emph{evaluation-driven adaptivity}.

%\subsection{Automatic Summary and Reporting}
%To make the system transparent and accessible to non-expert users, Data Agent automatically summarizes the entire decision and optimization process into an interpretable report.  During execution, all actions, decisions, metrics, and validation results are logged into a structured record.  At the end of the workflow, the system synthesizes these logs into a concise summary that explains: (i) which cleaning and feature operations were applied,  (ii) how routing decisions evolved,  (iii) what validation steps were triggered, and  (iv) how performance changed across iterations.   The summary is automatically generated in natural language and can be visualized through the demo interface.  This provides users with a clear narrative of how their raw data was transformed and why certain choices were made—allowing non-experts to understand, trust, and reuse the improved data without inspecting any code.  In this way, the summarization module completes the automation loop: it not only produces better data but also communicates the reasoning behind it.

% \vspace{-0.2cm}
\subsection{Automatic Summary and Reporting}
% \vspace{-0.1cm}
To ensure transparency and provenance, the system automatically logs the routing choices, all actions, grounding outcomes, machine learning performance metrics, and summarizes these logs into a structured, concise natural language report that outlines: 
(i) the cleaning and feature operations applied,  
(ii) the evolution of routing decisions,  
(iii) the validation checks performed, and  
(iv) the performance progression across iterations.  
The summary, viewable through the demo interface, translates internal reasoning into a human-readable narrative that lets non-experts trace data transformations and decisions, but also allows the feedback loop to collect experiences as benchmark data and perform self-reflection.

\subsection{Pseudo-code of \textsc{Dataforge}}

% \begin{algorithm}[htbp]
% \caption{\textsc{Dataforge}}
% \label{alg:dataforge}
% \small
% \begin{algorithmic}[1]
% \REQUIRE Dataset $\mathcal{D}$; budget $\mathcal{B}$ (MaxActions, TimeBudget, MinImprove)
% \ENSURE Best transformed data $\mathcal{D}^\star$ and pipeline $\pi^\star$
% \STATE $\mathcal{D}^\star \leftarrow \textsc{Clean}(\mathcal{D})$, $\pi^\star \leftarrow [\,]$, $u^\star \leftarrow -\infty$
% \FOR{$t = 1$ \TO MaxActions}
%   \IF{$\textsc{TimeExceeded}(\mathcal{B})$}
%     \STATE \textbf{break}
%   \ENDIF
%   \STATE $c \leftarrow \textsc{CoarseRoute}(\mathcal{D}^\star)$ \COMMENT{Selection/Transformation/Engineering}
%   \STATE $a \leftarrow \textsc{PlanAction}(c, \mathcal{D}^\star)$
%   \IF{$\neg\,\textsc{Ground}(a, \mathcal{D}^\star)$}
%     \STATE \textbf{continue}
%   \ENDIF
%   \STATE $\mathcal{D}' \leftarrow \textsc{Execute}(a, \mathcal{D}^\star)$
%   \STATE $u \leftarrow \textsc{Evaluate}(\mathcal{D}')$
%   \IF{$u > u^\star + \textsc{MinImprove}(\mathcal{B})$}
%     \STATE $\mathcal{D}^\star \leftarrow \mathcal{D}'$
%     \STATE $\pi^\star \leftarrow \pi^\star \circ [a]$
%     \STATE $u^\star \leftarrow u$
%   \ENDIF
%   \IF{$\textsc{AutoStop}(\mathcal{B})$}
%     \STATE \textbf{break}
%   \ENDIF
% \ENDFOR
% \RETURN $(\mathcal{D}^\star, \pi^\star)$
% \end{algorithmic}
% \end{algorithm}

\begin{algorithm}[htbp]
\caption{\textsc{Dataforge} (Compressed)}
\label{alg:dataforge}
\small
\begin{algorithmic}[1]
\REQUIRE Dataset $\mathcal{D}$; budget $\mathcal{B}$ (MaxActions, TimeBudget, MinImprove)
\ENSURE Best transformed data $\mathcal{D}^\star$ and pipeline $\pi^\star$
\STATE $\mathcal{D}^\star \leftarrow \textsc{Clean}(\mathcal{D})$; $\pi^\star \leftarrow [\,]$; $u^\star \leftarrow -\infty$
\FOR{$t=1$ \TO MaxActions}
  \IF{$\textsc{Stop}(\mathcal{B})$} \STATE \textbf{break} \ENDIF
  \STATE $c \leftarrow \textsc{CoarseRoute}(\mathcal{D}^\star)$ \COMMENT{Sel/Trans/Eng}
  \STATE $a \leftarrow \textsc{PlanAction}(c,\mathcal{D}^\star)$; \IF{$\neg\,\textsc{Ground}(a,\mathcal{D}^\star)$} \STATE \textbf{continue} \ENDIF
  \STATE $\mathcal{D}' \leftarrow \textsc{Execute}(a,\mathcal{D}^\star)$; $u \leftarrow \textsc{Evaluate}(\mathcal{D}')$
  \IF{$u > u^\star + \textsc{MinImprove}(\mathcal{B})$}
    \STATE $\mathcal{D}^\star \leftarrow \mathcal{D}'$; $\pi^\star \leftarrow \pi^\star \circ [a]$; $u^\star \leftarrow u$
  \ENDIF
\ENDFOR
\RETURN $(\mathcal{D}^\star,\pi^\star)$
\end{algorithmic}
\end{algorithm}

\textbf{Algorithm~\ref{alg:dataforge}} summarizes the overall workflow of \textsc{Dataforge}.
Starting from a cleaned dataset, the agent iteratively (i) performs coarse routing to choose a feature-operation family, (ii) plans a concrete action, (iii) grounds the action to ensure executability, and (iv) executes and evaluates the updated data.
The best-so-far pipeline is updated only when the downstream performance improves beyond a threshold, and the process terminates by budget or auto-stop.

\begin{table*}[htbp]
  \begin{center}
    \caption{
    Downstream performance comparison on regression and classification datasets.
    For regression, we report 1-RAE (Relative Absolute Error), and for classification, F1-score (\%).
    }
    \label{tab:downstream}
    \resizebox{\linewidth}{!}{
    \begin{tabular}{lcccccccccc}
        \toprule
        \textbf{Dataset} 
        & German Credit 
        & Amazon Employee 
        & Ionosphere 
        & PimaIndian 
        & Messidor Feature 
        & SVMGuide3 
        & OpenML 586 
        & OpenML 618 
        & Airfoil \\
        \midrule
        \textbf{Source} 
        & UCIrvine
        & Kaggle
        & UCIrvine
        & Kaggle
        & UCIrvine
        & LibSVM
        & OpenML
        & OpenML
        & UCIrvine \\
        \textbf{Task} 
        & Classification 
        & Classification 
        & Classification 
        & Classification 
        & Classification 
        & Classification 
        & Regression 
        & Regression 
        & Regression \\
        \textbf{Samples} 
        & 1{,}000 & 32{,}769 & 351 & 768 & 1{,}151 & 1{,}243 & 1{,}000 & 1{,}000 & 1{,}503 \\
        \textbf{Features} 
        & 24 & 9 & 34 & 8 & 19 & 21 & 25 & 50 & 5 \\
        \midrule
        \textbf{Original} 
        & 74.20\% & 93.37\% & 93.37\% & 80.68\% & 69.09\% & 81.85\% & 0.6311 & 0.4402 & 0.5749 \\
        \textbf{RDG} 
        & 68.01\% & 92.31\% & 91.17\% & 76.04\% & 62.38\% & 78.68\% & 0.5681 & 0.3720 & 0.5193 \\
        \textbf{LDA} 
        & 63.91\% & 91.64\% & 65.53\% & 63.80\% & 47.52\% & 65.24\% & 0.1109 & 0.0521 & 0.2201 \\
        \textbf{ERG} 
        & 74.43\% & 92.43\% & 92.02\% & 76.17\% & 66.90\% & 82.62\% & 0.6147 & 0.3561 & 0.5193 \\
        \textbf{NFS} 
        & 68.67\% & 93.21\% & 91.17\% & 74.87\% & 63.77\% & 79.16\% & 0.5443 & 0.3473 & 0.5193 \\
        \textbf{AFAT} 
        & 68.32\% & 92.97\% & 92.87\% & 76.56\% & 66.55\% & 79.49\% & 0.5435 & 0.2472 & 0.5210 \\
        \textbf{PCA} 
        & 67.92\% & 92.29\% & 92.87\% & 63.80\% & 67.21\% & 67.60\% & 0.1109 & 0.1016 & 0.2730 \\
        \textbf{TTG} 
        & 64.51\% & 92.79\% & 90.31\% & 74.48\% & 66.46\% & 79.81\% & 0.5443 & 0.3467 & 0.5003 \\
        \textbf{FeatLLM~\cite{han2024large}} 
        & 76.35\% & 93.62\% & 95.38\% & 89.66\% & 72.62\% & 81.17\% & 0.6477 & 0.4597 & 0.5587 \\
        \textbf{CAAFE~\cite{hollmann2023large}} 
        & 59.92\% & 91.41\% & 92.84\% & 79.86\% & 66.10\% & 81.74\% & N/A & N/A & N/A \\
        \textbf{AutoFeat~\cite{horn2019autofeat}} 
        & 74.86\% & 93.29\% & 93.37\% & 80.86\% & 69.08\% & 82.54\% & 0.6329 & 0.4407 & 0.5746 \\
        \textbf{OpenFE~\cite{zhang2023openfe}} 
        & 74.50\% & 93.44\% & 93.37\% & 80.86\% & 69.09\% & 83.05\% & 0.6311 & 0.4402 & 0.5746 \\
        \textbf{ELLM-FT~\cite{gong2025evolutionary}} 
        & 76.39\% & 93.17\% & 96.01\% & 89.66\% & 74.80\% & 82.70\% & 0.6328 & 0.4734 & 0.6174 \\
        \midrule
        \textbf{Pure LLM (Llama 3.1)} 
        & 75.43\% & 75.43\% & 96.68\% & 89.66\% & 75.61\% & 81.17\% & 0.7811 & 0.6653 & 0.6329 \\
        \textbf{RL-Policy-1 (GRFG~\cite{wang2022group})} 
        & 68.29\% & 68.29\% & 93.16\% & 75.39\% & 69.24\% & 79.81\% & 0.5768 & 0.4562 & 0.5587 \\
        \textbf{RL-Policy-2 (MOAT~\cite{wang2023reinforcement})} 
        & 72.44\% & 72.44\% & 95.69\% & 80.73\% & 73.02\% & 81.17\% & 0.6251 & 0.4734 & 0.5967 \\
        \textbf{\textsc{Dataforge}} 
        & \textbf{79.60\%} & \textbf{94.41\%} & \textbf{97.14\%} & \textbf{90.14\%} & \textbf{76.98\%} & \textbf{84.64\%} & \textbf{0.7849} & \textbf{0.7219} & \textbf{0.7492} \\
        \bottomrule
    \end{tabular}}
  \end{center}
\end{table*}

\section{Empirical Study}
We demonstrate the functionality and effectiveness of \textsc{Dataforge} through both quantitative evaluation and visual inspection of its autonomous workflow. The goal is to show that \textsc{Dataforge} achieves reliable performance across diverse datasets while providing interpretable and user-friendly automation for non-expert users.

\subsection{Setup}
\label{subsec:setup}

\textbf{Datasets and protocol.}
We evaluate on diverse tabular benchmarks from UCIrvine~\cite{uci_dataset_2023}, CPLM~\cite{CPLM_2023}, Kaggle~\cite{howard_kaggle_2023}, and OpenML~\cite{Openml_dataset_2023}, covering both classification and regression (statistics in \textbf{Table~\ref{tab:downstream}}).
All methods are evaluated with the same 5-fold cross-validation protocol, and we report the average score across folds.

\textbf{Downstream evaluator and metrics.}
We assess feature engineering utility by training a fixed downstream model on the dataset with identical hyperparameters for all methods.
For classification, we report macro-F1; for regression, we report $1$-RAE:
\[
1\text{-RAE} = 1 - \frac{\|\bm{y}_{pred} - \bm{y}_{real}\|_1}{\|\bm{y}_{real} - \bar{\bm{y}}_{real}\|_1}.
\]

\textbf{Operator set.}
All applicable methods share the same operator set $\mathcal{O}=\mathcal{O}_1 \cup \mathcal{O}_2$, where
$\mathcal{O}_1=\{\texttt{sqrt},\texttt{square},\texttt{cube},\texttt{reciprocal},\texttt{log},\\\texttt{sin},\texttt{cos},\texttt{tanh},\texttt{sigmoid},\texttt{standard},\texttt{normalize},\texttt{quantile}\}$
and $\mathcal{O}_2=\{+,-,\times,\div\}$.

\textbf{LLM backbones.}
We use Llama-3.2-3B as our base LLM for the agent.
We use a unified generation protocol with $\texttt{temperature}=0.7$, $\texttt{top\_p}=0.9$, and $\texttt{max\_new\_tokens}=500$.

\noindent\textbf{Baselines and implementation.}
We compare against: (i) no/linear baselines, including \textit{Original}, RDG, PCA~\cite{mackiewicz1993principal}, and LDA; 
(ii) classical automated feature engineering and search methods, including ERG, AFAT~\cite{horn2020autofeat}, AutoFeat~\cite{horn2019autofeat}, NFS~\cite{chen2019neural}, TTG~\cite{khurana2018feature}, and OpenFE~\cite{zhang2023openfe};
(iii) RL-based policies, including GRFG~\cite{wang2022group} and MOAT~\cite{wang2023reinforcement};
and (iv) LLM-based approaches, including FeatLLM~\cite{han2024large}, CAAFE~\cite{hollmann2023large}, ELLM-FT~\cite{gong2025evolutionary}, as well as a \textit{Pure LLM} variant.

\noindent\textbf{Configurations.}
All experiments were conducted on the Ubuntu 22.04.3 LTS operating system, with a 13th-generation Intel(R) Core(TM) i9-13900KF CPU and an NVIDIA GeForce RTX 4090 GPU. The experiments were conducted using Python 3.11.5 and PyTorch 2.0.1.

% \vspace{-0.2cm}
\subsection{Study of Overall Performances.}
% \vspace{-0.1cm}

\textbf{Table~\ref{tab:downstream}} reports the downstream utility of different feature-engineering approaches on six classification and three regression benchmarks.
We compare \textsc{Dataforge} with representative baselines from three families: classical automated feature engineering/search, RL-based policies, and LLM-based methods, as well as the \textit{Original} features.
The goal of this experiment is to evaluate whether an agentic, fully automated pipeline can consistently improve tabular representations across diverse datasets under a unified protocol.

Overall, \textsc{Dataforge} delivers the greatest and most consistent improvements across both classification and regression benchmarks.
Compared with classical automated feature engineering and search methods (e.g., AutoFeat, AFAT, NFS, TTG, and OpenFE), \textsc{Dataforge} more reliably discovers beneficial transformations, suggesting that iterative, feedback-driven planning is better suited for navigating large operator spaces than one-shot heuristics or fixed search strategies.
Compared with RL-based pipelines (GRFG/MOAT) and the \textit{Pure LLM} variant, \textsc{Dataforge} exhibits substantially higher stability: it avoids low-utility or invalid proposals and maintains strong performance across heterogeneous datasets with different sizes and feature dimensions.
These results indicate that coarse routing to an appropriate feature-operation family, followed by fine-grained action planning and grounding, effectively reduces candidate ambiguity and improves decision quality under a fixed budget.

\textbf{Insights:}
First, the consistent improvements over \textit{Original} features validate that \textsc{Dataforge} can autonomously refine tabular representations via a combination of selection, transformation, and feature construction, rather than relying on a single operator family.
Second, the performance advantages are achieved without training a task-specific policy (unlike RL methods), making the approach readily deployable across datasets.
Finally, \textsc{Dataforge} requires minimal user input (dataset and basic runtime constraints) while producing a fully automated pipeline, supporting the practicality of agentic data preparation as a general paradigm for tabular feature engineering.

\subsection{Ablation Study}

\textbf{Table~\ref{tab:ablation}} ablates key components of \textsc{Dataforge} on three representative datasets (German Credit, Amazon Employee, and OpenML~618).
For each variant, we report the downstream score (macro-F1 for classification and 1-RAE for regression), the valid-action rate Valid(\%) (fraction of proposed actions that pass grounding and execute successfully), and the average number of evaluator calls, which reflects the interaction cost of the iterative pipeline.

\begin{table}[htbp]
\centering
\caption{Ablation study of \textsc{Dataforge} on three representative datasets.
Score is macro-F1 (\%) for classification and 1-RAE for regression.
Valid(\%) reports the ratio of proposed actions that pass grounding and are executed successfully; Calls denotes the average number of evaluator calls.}
\label{tab:ablation}
\resizebox{\linewidth}{!}{
\begin{tabular}{lccc|ccc|ccc}
\toprule
& \multicolumn{3}{c}{German Credit (C)} 
& \multicolumn{3}{c}{Amazon Employee (C)} 
& \multicolumn{3}{c}{OpenML 618 (R)} \\
\cmidrule(lr){2-4}\cmidrule(lr){5-7}\cmidrule(lr){8-10}
\textbf{Variant}
& \textbf{Score} & \textbf{Valid(\%)} & \textbf{Calls}
& \textbf{Score} & \textbf{Valid(\%)} & \textbf{Calls}
& \textbf{Score} & \textbf{Valid(\%)} & \textbf{Calls} \\
\midrule
Full (\textsc{Dataforge}) 
& 79.60\% & 98.51\% & 3
& 94.41\% & 99.90\% & 6
& 0.7219 & 98.84\% & 4 \\
w/o Routing 
& 76.86\% & 97.01\% & 2
& 93.96\% & 98.33\% & 3
& 0.7169 & 90.41\% & 3 \\
w/o Grounding 
& 77.49\% & 91.60\% & 5
& 94.36\% & 84.10\% & 11
& 0.7031 & 75.76\% & 5 \\
w/o Perf Loop 
& 76.27\% & 96.30\% & 1
& 93.95\% & 97.00\% & 1
& 0.6653 & 98.25\% & 1 \\
\bottomrule
\end{tabular}}
\end{table}

The full system achieves the best overall scores while maintaining high Valid(\%) and modest call counts.
Removing grounding (\textit{w/o Grounding}) leads to similar or slightly lower scores but a substantial drop in Valid(\%) (e.g., to 84.10\% on Amazon Employee and 75.76\% on OpenML~618) and a clear increase in calls (e.g., 6$\rightarrow$11 on Amazon Employee), indicating frequent invalid attempts and repeated retries.
Without routing (\textit{w/o Routing}), both performance and Valid(\%) degrade (especially on OpenML~618), while calls are sometimes smaller; this is consistent with unstructured candidate selection causing early termination when sampled actions fail to improve over the current best state.
Finally, without the performance loop (\textit{w/o Perf Loop}), the system degenerates into a one-shot variant with Calls$=1$, which yields notably worse scores, particularly on regression (0.7219$\rightarrow$0.6653).

\textbf{Insights:}
These ablations highlight complementary roles of the components.
Grounding primarily improves \emph{reliability} by filtering non-executable actions, thereby reducing wasted attempts even when the final score is not dramatically affected.
Hierarchical routing improves \emph{decision quality} by narrowing the candidate space; without it, the agent is more likely to try low-utility actions and stop early under the same auto-stop rule.
The performance optimization loop is crucial for \emph{refinement}: iterative evaluation and best-so-far updates are needed to move beyond a first valid action and consistently reach the strongest pipeline under a fixed budget.

\subsection{Study of Reliability and Automation.} 

We evaluate whether \textsc{Dataforge} can run \emph{end-to-end autonomously} and remain reliable under iterative feature operations.
In this study, we focus on two practical indicators: (i) \emph{execution reliability}, measured by the rate of invalid attempts during action proposal and execution, and (ii) \emph{automation stability}, reflected by the number of retries needed to obtain valid actions and complete a run.

\begin{figure}[tbp]
  \centering
  \begin{subfigure}[t]{0.48\linewidth}
    \centering
    \includegraphics[width=\linewidth]{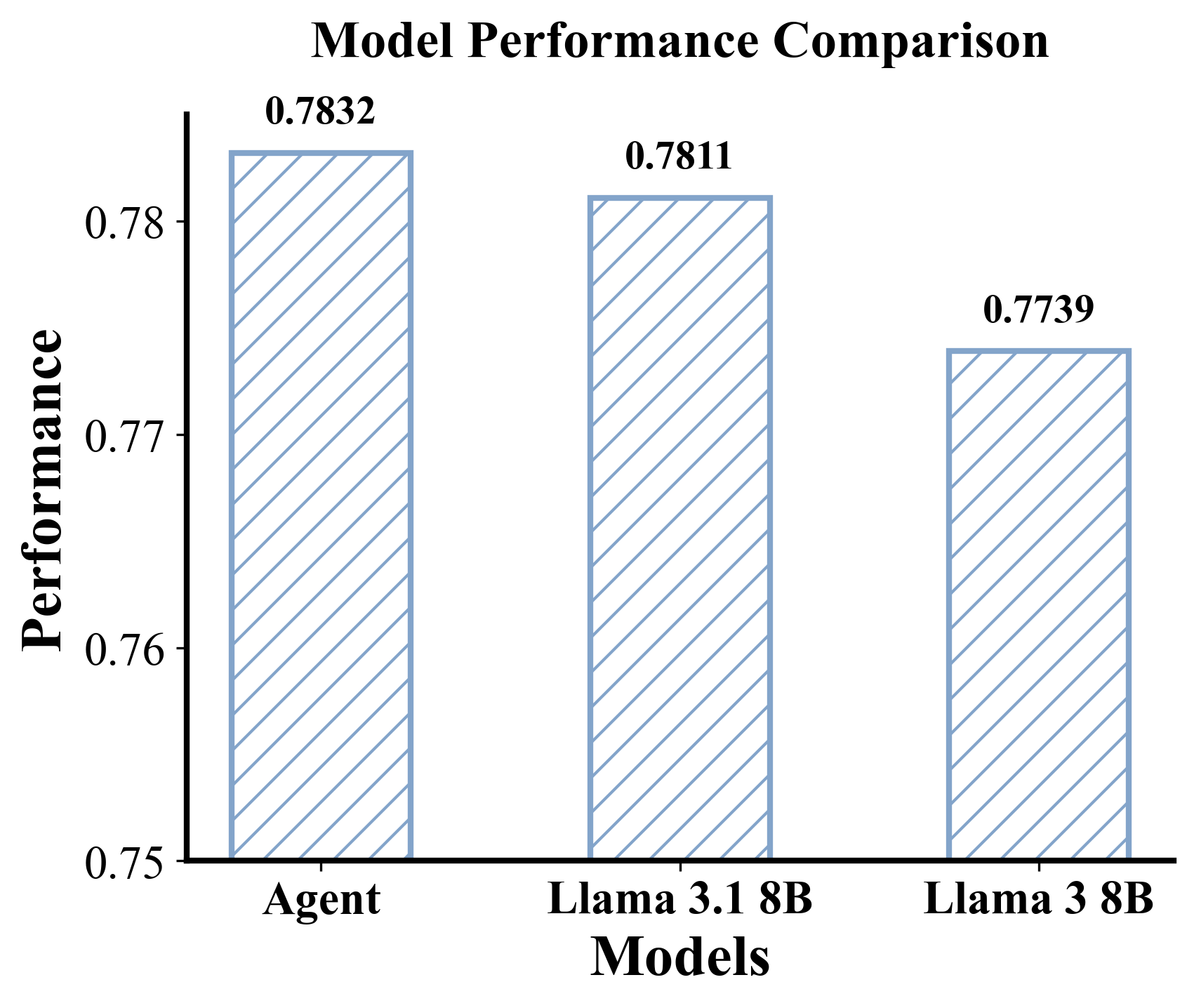}
    \caption{F1-score comparison.}
    \label{fig:autonomous_f1}
  \end{subfigure}
  \hfill
  \begin{subfigure}[t]{0.48\linewidth}
    \centering
    \includegraphics[width=\linewidth]{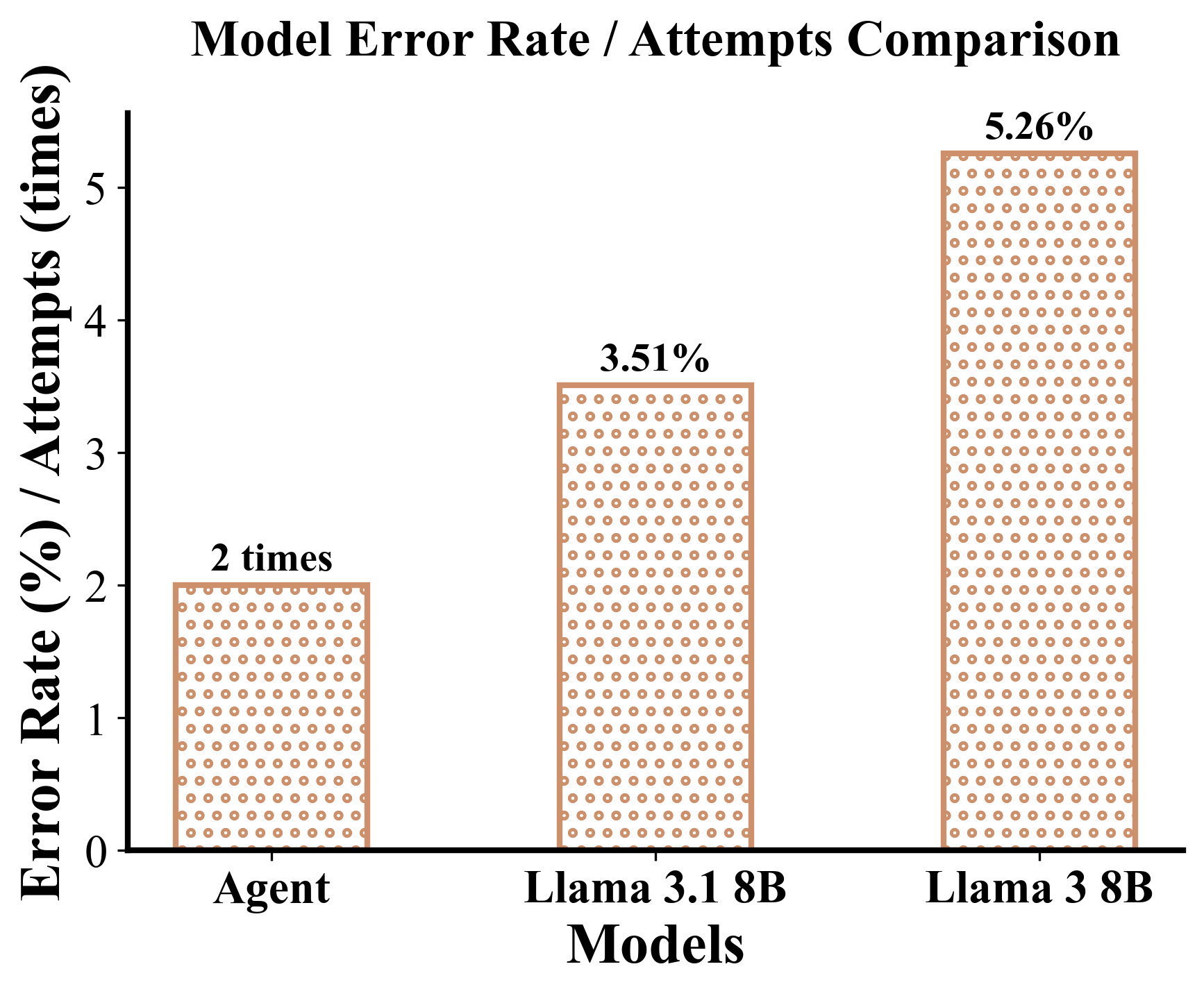}
    \caption{Error rate comparison.}
    \label{fig:autonomous_error}
  \end{subfigure}
  \caption{Comparison of modular agent (DataAgent) and pure LLM baselines, showing higher reliability and fewer retries for the modular agent paradigm.}
  \label{fig:autonomous}
\end{figure}

As shown in \textbf{Figure~\ref{fig:autonomous}}, \textsc{Dataforge} exhibits consistently higher reliability than pure LLM baselines.
Across all nine datasets, it completes the entire pipeline without interruption and typically converges with about two action retries on average.
In contrast, directly using Llama~3.1~8B or Llama~3~8B to propose feature operations leads to a non-trivial failure rate (around 3--5\%), indicating that purely generative pipelines are more likely to produce invalid or non-executable operations.
Meanwhile, \textsc{Dataforge} achieves comparable or slightly better downstream utility (average score $\approx 0.783$), suggesting that improved reliability does not come at the expense of performance.

\textbf{Insights:}
These results highlight the importance of the modular agent design.
By combining hierarchical routing with grounding validation, \textsc{Dataforge} constrains the action space and filters out unsafe proposals before execution, thereby reducing retries and preventing workflow breakdowns.
This makes the system more predictable for continuous operation and supports the goal of \emph{fully automated} data preparation with minimal user intervention.

\subsection{Study of Efficiency Trade-offs.} 
We further study the efficiency--performance trade-offs of \textsc{Dataforge} by comparing it with RL-based pipelines that explicitly optimize feature operations via training.
We report three aspects of efficiency in \textbf{Table~\ref{tab:efficiency}}: \emph{training cost} (wall-clock training time), \emph{inference latency} (time to complete a single run), and \emph{interaction cost} measured by the number of calls (i.e., the number of evaluator/tool invocations needed to obtain a final result).
This experiment aims to characterize whether an agentic pipeline can remain responsive while still achieving strong downstream utility.

\begin{table}[h]
  \begin{center}
    \caption{Efficiency comparison: time (seconds) and call times.}
    \label{tab:efficiency}
    \resizebox{\linewidth}{!}{\begin{tabular}{lccc}
        \toprule
        Metric & \makecell{RL-Policy-1 \\ (GRFG~\cite{wang2022group})} & \makecell{RL-Policy-2 \\ (MOAT~\cite{wang2023reinforcement})} & \makecell{\textsc{Dataforge}} \\
        \midrule
        Training Time (sec) & 1493.7 & 649.4 & 0 \\
        Inference Time (sec) & 0.5 & 2.14 & 3.89 \\
        Calls & 13 & 22 & 2 \\
        Performance & 0.5768 & 0.6251 & \textbf{0.7832} \\
        \bottomrule
    \end{tabular}}
  \end{center}
\end{table}

\textbf{Table~\ref{tab:efficiency}} shows that \textsc{Dataforge} eliminates the dominant overhead of RL-style optimization.
Both RL-Policy-1 (GRFG) and RL-Policy-2 (MOAT) require substantial training time (1493.7\,s and 649.4\,s), whereas \textsc{Dataforge} shows zero training cost and can be directly applied to new datasets.
At inference time, \textsc{Dataforge} completes a full workflow in 3.89\,s on average, which is slightly slower than RL inference (0.5--2.14\,s) but remains lightweight in absolute terms.
Notably, it achieves this with only two calls on average, compared to 13--22 calls for the RL baselines, indicating a markedly lower interaction/sample complexity under the same evaluation setting.
Despite the reduced call budget, \textsc{Dataforge} attains the best downstream performance (0.7832), suggesting that its iterative planning is highly cost-effective.

\textbf{Insights:}
These results highlight a favorable operating point for practical deployment.
While RL policies can achieve fast inference after training, the up-front training time and higher call counts make them less attractive when datasets change frequently or when rapid, on-demand data preparation is required.
In contrast, \textsc{Dataforge} trades a small increase in per-run latency for eliminating training altogether and substantially reducing interaction cost, yielding strong performance with high responsiveness.
This supports the use of modular agentic feature engineering as a practical alternative to training-heavy optimization methods.

% \subsection{Takeaways on Paradigms for Tabular Data}
% From these comparisons, several insights emerge:

% \paragraph{(1) Classical and AutoML baselines.}  
% These methods are mature and stable but depend on fixed model families and handcrafted transformations, limiting adaptability to diverse datasets.

% \paragraph{(2) Pure LLM approaches.}  
% LLMs offer automation and flexibility but lack grounding, often producing invalid outputs or silent failures, making them less reliable as standalone systems.

% \paragraph{(3) RL-based policies.}  
% Reinforcement learning can achieve strong results but requires heavy, dataset-specific training and environment feedback, reducing practicality.

% \paragraph{(4) Modular agent paradigm (Dataforge).}  
% By integrating rule-based and LLM-based routing with grounding and dual feedback loops, Dataforge balances autonomy and safety.  
% It adapts to new datasets in a zero-shot manner, ensuring end-to-end automation with interpretable outcomes.  
% These results demonstrate the feasibility of autonomous and reliable data transformation, marking a concrete step toward practical \textit{Dataforges} in real-world AI pipelines.

\section{Demo and Case Studies}

% \vspace{-0.2cm}
\subsection{User Interface}
% \vspace{-0.1cm}

\begin{figure}[htbp]
  \centering
  \includegraphics[width=\linewidth]{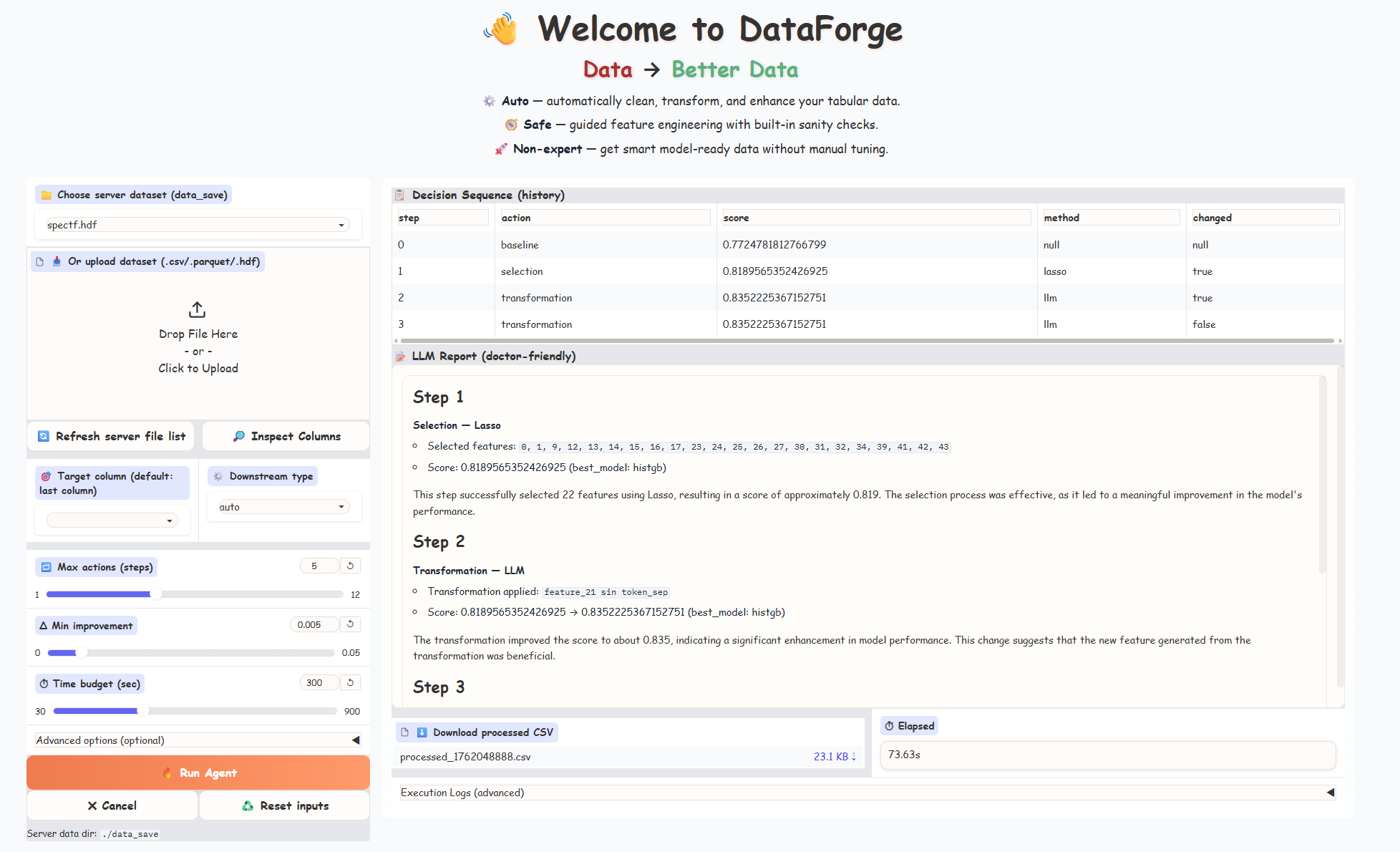}
  \caption{The interface of \textsc{Dataforge}.}
  \label{fig:interface}
\end{figure}

\textbf{Figure.~\ref{fig:interface}} illustrates the user interface of \textsc{Dataforge}, which offers an intuitive environment for autonomous tabular data transformation.  
Users can upload datasets in standard formats (\texttt{.csv}, \texttt{.parquet}, and \texttt{.hdf}), label predictors and response column in a machine learning task, and configure runtime parameters, such as, \textit{Max Actions}, \textit{Min Improvement}, and \textit{Time Budget}.  
After uploading datasets, users can select from pre-loaded examples. 
During execution, the interface visualizes real-time logs of thinking steps in reasoning, data cleaning and feature engineering actions,  and predictive performance metrics (e.g.,  F1-score, AUC, MAE, and RMSE depending on classification, regression, and unsupervised tasks ), so that users can observe how the data agent progressively refines data quality. 
Upon completion, it summarizes all executed actions, validation outcomes, and final performance statistics, to provide transparency and provenance, as well as collect action historical trajectories for benchmark data of training data agents.  
The interface is designed to be modular and extensible, supporting future integration with interactive dashboards and collaborative agent environments.

% \vspace{-0.2cm}
\subsection{Toward Accurate and Efficient Heart-Disease Detection}
% \vspace{-0.1cm}

Accurate identification of heart disease is one of the central challenges in clinical cardiology. 
Modern diagnostic models rely on dozens of imaging and physiological indicators, but these measurements are often noisy, redundant, or inconsistent across patients. 
As a result, even experienced cardiologists face difficulty in building reliable predictive models from raw data, and existing computational tools usually require manual data engineering and expert parameter tuning.

\begin{figure}[htbp]
  \centering
  \includegraphics[width=\linewidth]{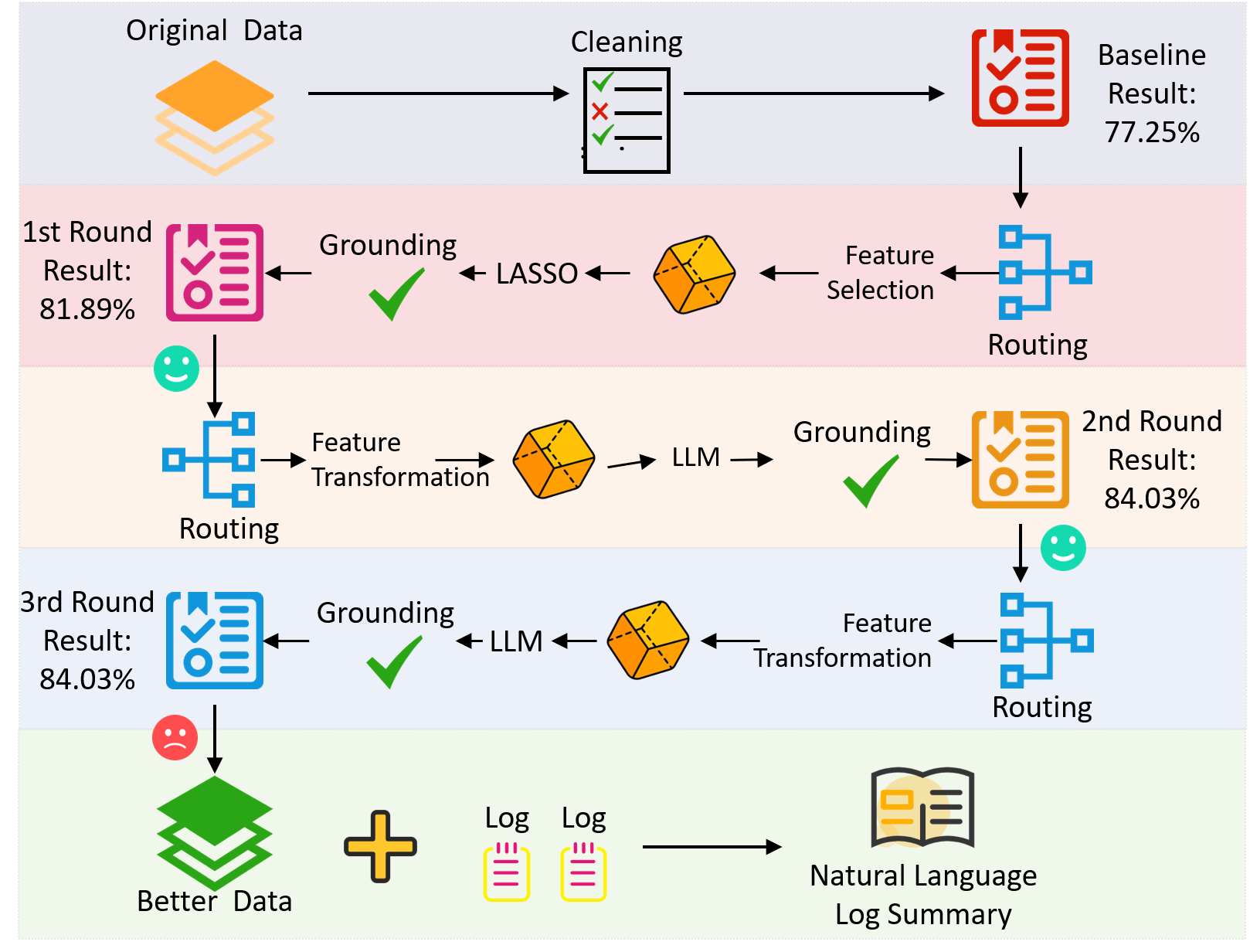}
  \caption{\textsc{Dataforge} Dealing with Heart-Disease Detection.}
  \label{fig:case}
\end{figure}

To address this challenge, \textsc{DataForge} is applied to an automated diagnostic classification task using the SPECTF Heart dataset, which contains patient-level imaging signals used for detecting abnormal cardiac function. 
The system autonomously analyzes the uploaded dataset, identifies the most informative diagnostic indicators, and removes irrelevant or conflicting variables. 
It then constructs refined combinations of cardiac features that better represent patient conditions, iteratively evaluating improvements until no further diagnostic gain can be achieved.  

Throughout this process (\textbf{Figure.~\ref{fig:case}}), no manual intervention or algorithmic knowledge is needed. 
The workflow completes in about 100 seconds and generates a concise report and a new dataset that achieves higher diagnostic accuracy with fewer features, reducing the total number of variables from 44 to 20 while improving the predictive score from 0.772 to 0.840. 
This demonstrates that \textsc{DataForge} can serve as an intelligent assistant for cardiology, automatically transforming complex measurements into a cleaner, smaller, and more reliable dataset for accurate and efficient heart-disease detection.

\subsection{Chemical Design}

\begin{figure}[htbp]
    \centering
    \includegraphics[width=0.9\linewidth]{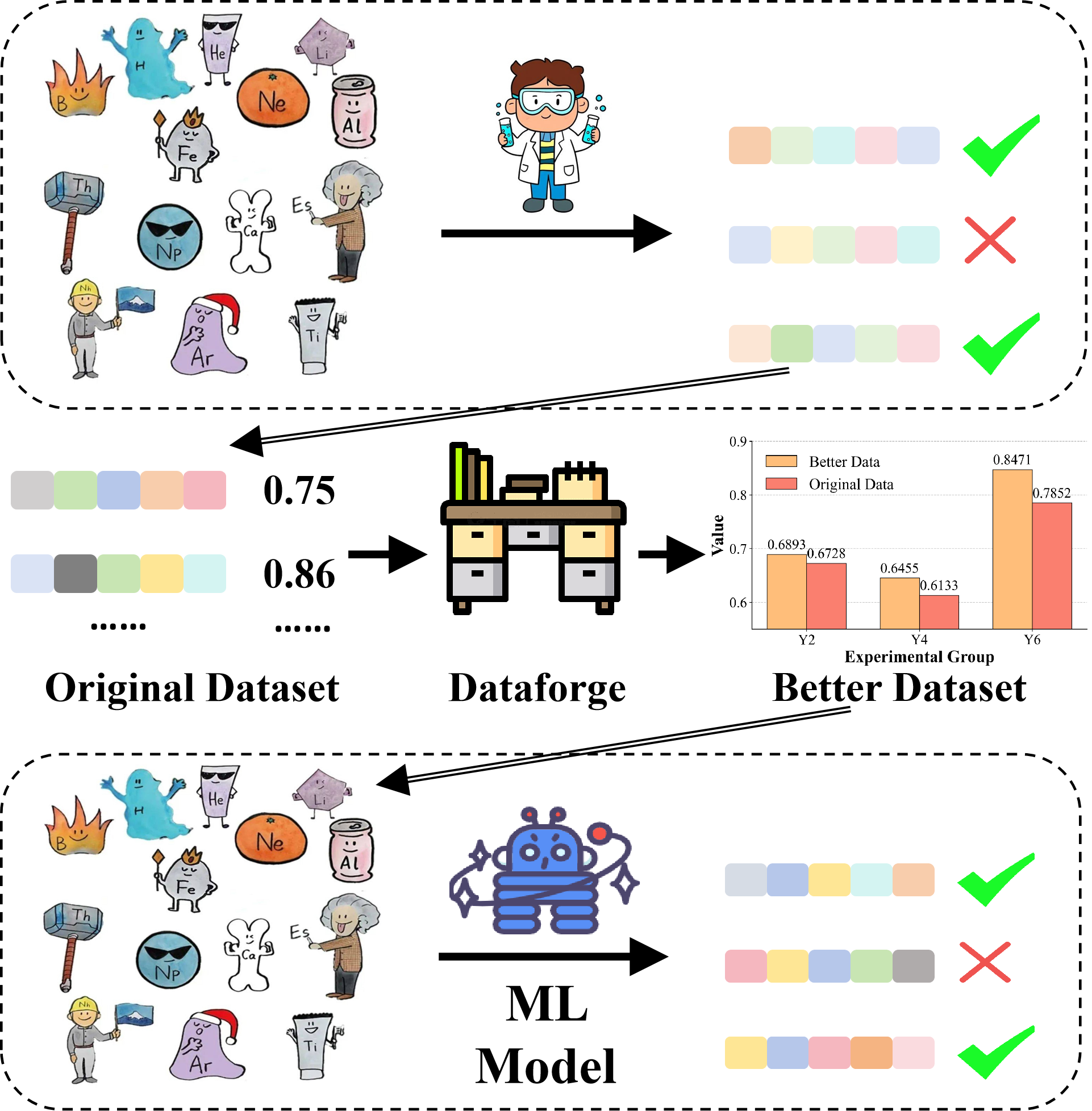}
    \caption{\textbf{\textsc{Dataforge} for materials discovery.}}
    \label{fig:chem_overview}
\end{figure}

\textbf{Figure~\ref{fig:chem_overview}} illustrates how \textsc{Dataforge} supports a practical materials discovery workflow.
Starting from a small seed set of physical experiments, we obtain an initial tabular dataset with multiple target properties.
\textsc{Dataforge} then enhances the dataset via automated feature operations and produces an AI-ready version together with traceable logs and a natural-language report.
Training a downstream property predictor on the enhanced data improves predictive utility across targets, which enables more reliable screening and prioritization of candidate formulations, thereby reducing the number of costly follow-up experiments.

\section{Related Work}

% Feature Transformation (FT) aims to improve the feature space by applying mathematical operations to original features, so as to enhance the performance of downstream models.
% A major challenge lies in the combinatorial explosion of feature--operator compositions, which makes exhaustive enumeration infeasible and motivates automated feature engineering (AutoFE) methods.

\subsection{Automated Feature Engineering}

\textbf{Feature transformation.}
Prior work formulates feature transformation as searching over composition trajectories of operators and operands.
Early systems suggest generating a large candidate pool followed by selection to control redundancy~\cite{katz2016explorekit}.
To improve trajectory quality, many methods adopt structured discrete search, including heuristic-guided exploration and other discrete optimization strategies~\cite{kanter2015discrete1,khurana2016discrete2,tran2016discrete3}, as well as evolutionary algorithms for population-based exploration~\cite{zhu2022evolutionary1,gong2024evolutionary}.
Reinforcement learning further models transformations as sequential decision making and learns policies from downstream rewards~\cite{wang2022group,ying2025distribution,bai2025privacy}, with traceable/evolutionary variants emphasizing controllable search dynamics and interpretability~\cite{xiao2023traceable,xiao2024evolutionary2}.
Beyond discrete search, continuous optimization methods embed feature programs into differentiable latent spaces and apply gradient-based updates~\cite{wang2023reinforcement,zhu2022difer,gong2025sculpting}, while facing the challenge of mapping optimized embeddings back to valid executable transformations.

\textbf{Feature selection.}
Classical feature selection includes similarity/spectral criteria (e.g., Laplacian Score~\cite{he2005laplacian}, SPEC~\cite{zhao2007spectral}, Fisher Score~\cite{duda2006pattern}, Trace Ratio~\cite{nie2010efficient}), information-theoretic filters (e.g., Information Gain~\cite{lewis1992feature}, mutual information~\cite{battiti1994using}, mRMR~\cite{peng2005feature}, FCBF~\cite{yu2003feature}), sparse learning (e.g., $\ell_p$-regularized methods~\cite{tibshirani1996regression,zhu20031}), and statistics-based measures (e.g., T-test~\cite{davis1986statistics}, Chi-square~\cite{liu1995chi2}, Gini~\cite{gini1912variability}, CFS~\cite{hall1999feature}).
Recent work leverages deep and RL formulations to improve scalability and search quality, including deep feature selection~\cite{xiao2023beyond,ying2024feature,ying2024revolutionizing}, RL-based selection~\cite{wang2025knockoff}, Monte-Carlo style methods (MARFS~\cite{liu2023interactive}, MCRFS~\cite{liu2021efficient}), and specialized settings such as incomplete multi-view unsupervised selection~\cite{huang2023imufs}.

\textbf{LLMs for feature engineering.}
Recent studies explore using LLMs to propose feature transformations or engineered features via natural-language reasoning, enabling broader operator coverage and reduced manual effort~\cite{han2024large,hollmann2023large,gong2025evolutionary,gong2025unsupervised,gong2025agentic,wang2025llm, xie2025transformer}.
However, purely generative pipelines can suggest invalid or non-executable actions and often require additional validation and iterative control to ensure reliability and stable improvements.

\subsection{Agentic Platform}
Agentic systems augment LLMs with multi-step reasoning, planning, and tool use~\cite{zheng2026towards,wang2025mllm, xie2026agent}, enabling them to iteratively act in an environment rather than producing one-shot outputs~\cite{yin2025livemcp}.
A representative paradigm is ReAct, which interleaves reasoning traces with action/tool calls to improve task completion and error recovery~\cite{yao2022react}.

Building on general tool-using agents, web agents study how LLMs perceive and interact with complex web environments (e.g., multi-page, multi-tab interfaces) to accomplish real-world browsing tasks.
Benchmarks such as WebArena provide realistic web interaction environments for evaluating such agents~\cite{zhou2023webarena}, and systems like WebVoyager demonstrate end-to-end web navigation and action execution at scale~\cite{he2024webvoyager}.

Recently, the same agentic principles have been extended to the data domain.
Fu et al.\ propose \emph{autonomous data agents} that combine task decomposition, action reasoning, grounding into executable code/tool calls, and adaptive workflow planning to automate data-to-knowledge pipelines~\cite{fu2025autonomous}.
This perspective highlights new opportunities and challenges for trustworthy, scalable data operations (e.g., preprocessing, selection, transformation, and repairs) and motivates agentic platforms like ours that emphasize grounded execution and iterative feedback for reliable data engineering~\cite{fu2025autonomous}.

% \section{Conclusion and Future Work}
% \textcolor{red}{new}
% This demo presents \textit{Data Agent}, a fully autonomous and safe system for tabular data transformation. 
% By integrating hierarchical routing, dual feedback loops, and automatic summarization, the system realizes the ``From Data to Better Data'' pipeline—turning raw data into cleaner, more informative, and AI-ready datasets without human intervention. 
% The experiments demonstrate that Data Agent achieves reliable performance and stable automation across diverse tasks, highlighting the potential of modular agent architectures for practical data-centric AI.

% Future work will focus on extending the current prototype beyond tabular data to support multimodal inputs such as text, time series, and sensor data. 
% We also plan to enhance the agent’s memory and experience replay mechanisms for continual learning, and to establish standardized benchmarks for evaluating autonomous data agents. 
% Ultimately, our goal is to advance Data Agent toward a general-purpose, trustworthy, and interpretable data intelligence framework that can empower non-experts to leverage data effectively.

\section{Conclusion}
We presented \textsc{Dataforge}, an LLM-powered agentic data engineering platform for tabular data. It is \emph{automatic}, \emph{safe}, and \emph{non-expert friendly}.
By combining hierarchical routing with evaluation-driven refinement and automatic stopping, \textsc{Dataforge} transforms raw datasets into AI-ready representations end-to-end.
Experiments on nine public benchmarks show strong and robust gains over classical AutoFE/search, RL-based, and LLM baselines, while ablations highlight the roles of routing, grounding, and iterative optimization.

% In future work, we will extend \textsc{Dataforge} to richer data modalities and domain-specific operators, strengthen trust and safety with stricter constraints and uncertainty-aware control, and scale experience reuse to improve generalization across datasets and tasks.

% \begin{acks}
% \end{acks}

\bibliographystyle{ACM-Reference-Format}
\bibliography{sample-base}

% \input{sec/11-appendix}

\end{document}